\documentclass[amsmath, amssymb, aps, pra, twocolumn,floatfix,nofootinbib,superscriptaddress, showkeys]{revtex4}
\usepackage[utf8]{inputenc}
\usepackage[urlcolor=blue, colorlinks=true,citecolor=blue]{hyperref}
\usepackage{enumitem}

\usepackage{amsmath, amssymb}
\usepackage{textcomp, gensymb}
\usepackage{tabularx}
\usepackage{siunitx}

\usepackage{wrapfig}
\usepackage{graphicx}
\usepackage{float}
\usepackage{subfigure}
\usepackage{amssymb}
\usepackage{bbold}
\usepackage{color}
\usepackage{ulem}
\usepackage{qcircuit}

\newcommand{\ket}[1]{\left|#1\right\rangle}

\begin{document}

\title{Photovoltaic power forecasting using quantum machine learning}

\author{Asel Sagingalieva}
\affiliation{Terra Quantum AG, 9000 St. Gallen, Switzerland}

\author{Stefan Komornyik}
\affiliation{HAKOM Time Series GmbH, 1230 Vienna, Austria}

\author{Arsenii Senokosov}
\affiliation{Terra Quantum AG, 9000 St. Gallen, Switzerland}

\author{Ayush Joshi}
\affiliation{Terra Quantum AG, 9000 St. Gallen, Switzerland}

 \author{Christopher~Mansell}
 \affiliation{Terra Quantum AG, 9000 St. Gallen, Switzerland}

\author{Olga Tsurkan}
\affiliation{Terra Quantum AG, 9000 St. Gallen, Switzerland}

\author{Karan Pinto}
\affiliation{Terra Quantum AG, 9000 St. Gallen, Switzerland}

\author{Markus Pflitsch}
\affiliation{Terra Quantum AG, 9000 St. Gallen, Switzerland}

\author{Alexey Melnikov}
\thanks{Corresponding author, e-mail: alexey@melnikov.info
\begin{center}
\fbox{
\begin{minipage}{0.45\textwidth}
Please check the published version, which includes all the latest additions and corrections: Sol. Energy 302:114016, 2025, DOI: \href{https://doi.org/10.1016/j.solener.2025.114016}{10.1016/j.solener.2025.114016}
\end{minipage}
}
\end{center}
}
\affiliation{Terra Quantum AG, 9000 St. Gallen, Switzerland}

\begin{abstract}
Accurate forecasting of photovoltaic power is essential for reliable grid integration, yet remains difficult due to highly variable irradiance, complex meteorological drivers, site geography, and device-specific behavior. Although contemporary machine learning has achieved successes, it is not clear that these approaches are optimal: new model classes may further enhance performance and data efficiency. We investigate hybrid quantum neural networks for time-series forecasting of photovoltaic power and introduce two architectures. The first, a Hybrid Quantum Long Short-Term Memory model, reduces mean absolute error and mean squared error by more than $40\%$ relative to the strongest baselines evaluated. The second, a Hybrid Quantum Sequence-to-Sequence model, once trained, it predicts power for arbitrary forecast horizons without requiring prior meteorological inputs and achieves a $16\%$ lower mean absolute error than the best baseline on this task. Both hybrid models maintain superior accuracy when training data are limited, indicating improved data efficiency. These results show that hybrid quantum models address key challenges in photovoltaic power forecasting and offer a practical route to more reliable, data-efficient energy predictions.
\end{abstract}

\keywords{quantum machine learning, hybrid quantum neural network, quantum depth-infused layer, quantum LSTM, solar energy, photovoltaic power, time series}

\maketitle

\section{Introduction}


Electricity generation prediction, especially for photovoltaic (PV) systems, is a crucial tool for renewable energy adoption~\cite{sabri2022novel, li2020review}. 
The global economy must radically reduce emissions to stay within the $1.5\degree$C pathway (Paris Agreement) and the transition to renewable energy sources is necessary to achieve these objectives~\cite{ellabban2014renewable}. 
According to the International Energy Agency, solar PV’s installed power capacity is poised to surpass that of coal by 2027, becoming the world's largest power source.

Accurate PV power forecasts are vital for multiple facets of the energy industry such as long-term investment planning, regulatory compliance for avoiding penalties, and renewable energy management across storage, transmission, and distribution activities. 
Several studies show that an increase in forecasting accuracy reduces electricity generation from conventional sources. 
Increased accuracy also reduces operating costs of systems by reducing the uncertainty of PV power generation~\cite{anido2016value}. 
Such forecasts support improving the stability and sustainability of the power grid through optimizing power flow and counteracting solar power's intermittent nature~\cite{lee2021pv}. 
Such predictions are foundational in increasing the economic viability and improving the adoption of solar energy as they inform pricing and economic dispatch strategies, bolster competitiveness and over time reduce reliance on reserve power. 
Additionally, they assist with effective energy-storage management and the integration of PV systems into the power grid~\cite{bozorgavari2019two}, which is essential for the enduring success of renewable energy solutions~\cite{diagne2013review}.


Traditional methods for predicting PV power have primarily relied on statistical models, machine learning algorithms, or a blend of both~\cite{ahmed2019review}.
In Ref.~\cite{sabri2022novel}, a hybrid GRU–CNN model is proposed for forecasting the output of the DKASC 1B PV system (Alice Springs), showing improvements across all metrics compared with standalone GRU/LSTM/CNN models.
In Ref.~\cite{gomez2020photovoltaic}, the authors test an MLP in three scenarios for a plant in Monteroni di Lecce (2012–2013) and obtain nRMSE values of $2.94–9.98\%$.
Another study~\cite{kaloop2021novel} involves the training ANFIS-based hybrid algorithms for the University of Salento PV plant, achieving a minimum RMSE of $0.081$.
Applying kNN to the University of Salento data produced an absolute error of $5.5\%$.
In Ref.~\cite{shi2012forecasting}, the time series of a grid-connected PV plant in southern China are forecast using SVM models, yielding an average MRE of $8.64\%$.
A recent study implements a hybrid “FFNN + SVM” scheme optimized with particle swarm optimization (PSO) for long-term forecasting using real climate and load data; it achieves very high accuracy (correlation coefficient $R^2\approx0.9984$) but notes that random variability in the inputs can slow model convergence~\cite{boum2022photovoltaic}.
Another work proposes a deep-learning ensemble—an MLP combined with an LSTM—which outperforms classical approaches on demand and generation datasets~\cite{mbey2024novel}; in a subsequent study, a multi-objective PSO is integrated into an FFNN–LSTM hybrid, further reducing errors (MAPE $< 0.1\%$) for joint PV and load forecasting at the cost of increased model complexity~\cite{mbey2024solar}.
In addition, an ensemble of SVM and GRU models optimized by an ant colony algorithm aggregates their outputs and captures weather-induced variability across multiple PV sites, achieving a high level of agreement (correlation coefficient $R^2\approx0.9986$)~\cite{souhe2024optimized}, though the authors emphasize the need for more advanced fusion strategies and real-time adaptation for practical deployment.
Nevertheless, the variable and nonlinear nature of solar generation, influenced by a wide range of meteorological factors, remains a serious challenge for traditional models~\cite{blaga2019current}.

In light of these challenges, quantum machine learning (QML) emerges as a promising avenue. 
This rapidly evolving field, which melds the principles of quantum mechanics with classical machine learning~\cite{Schuld_2014, biamonte2017quantum, Dunjko.Briegel.2018, qml_review_2023}, can offer enhanced capabilities for improving the forecasting accuracy of time series tasks~\cite{emmanoulopoulos2022quantum}, including PV power generation~\cite{sushmit2023forecasting}. 
QML's potential arises from quantum features like superposition and entanglement, promising exponential speedups in certain tasks~\cite{schuld2018implementing}. 
Moreover, QML algorithms produce inherently probabilistic results, aptly suited for prediction tasks, and they may potentially function within an exponentially larger search space, amplifying their efficacy~\cite{Lloyd2013, 365700, Lloyd1996UniversalQS, schuld2020circuit}. 
Nonetheless, implementing quantum algorithms bears its own set of challenges, such as the need for error correction and sensitivity to external interference~\cite{Aaronson2015}. 
Yet, in spite of these challenges, hybrid quantum-classical models, especially hybrid quantum neural networks (HQNNs), have showcased their potential in diverse industrial realms, including healthcare~\cite{jain2022hybrid, pharma, steatosis_sagingalieva, sagingalieva2025hybrid, anoshin2024hybrid}, energy~\cite{kurkin2025forecasting, sushmit2023forecasting, lee2025predictive}, aerospace~\cite{rainjonneau2023quantum}, logistics~\cite{haboury2023supervised} and, automotive~\cite{sagingalieva2023hyperparameter}.

\begin{figure*}[ht!]
    \includegraphics[width=1\linewidth]{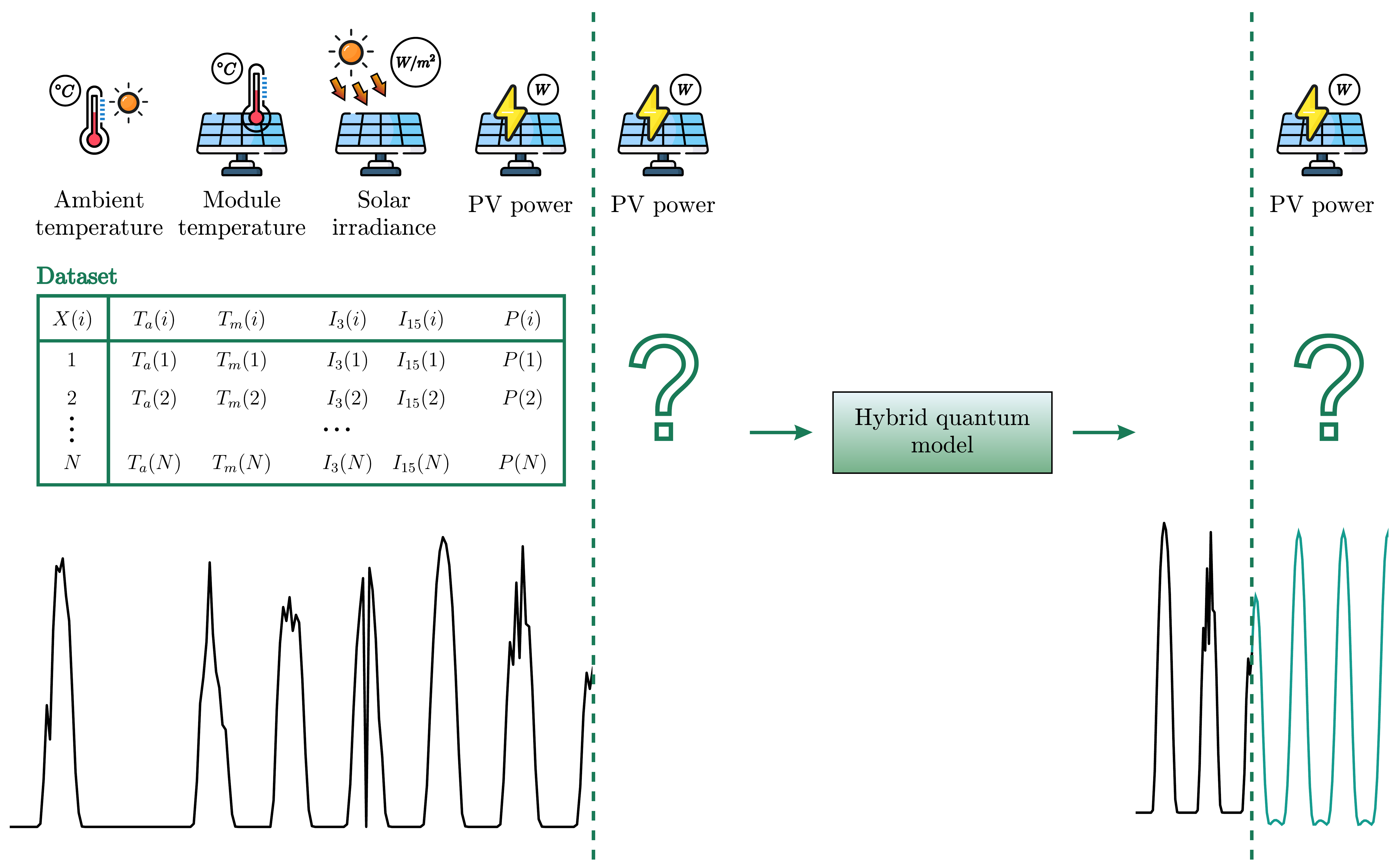}
    \caption{The input to a hybrid quantum model is presented as a chronological data table, documenting hourly meteorological parameters including ambient temperature ($T_a$), module temperature ($T_m$), and solar irradiance ($I_3$, $I_{15}$), alongside the mean PV power output ($P$). 
    The model is designed to leverage these data to generate predictions of PV power output for a short-term forecast aimed at near-future output, typically the next hour, and a long-term forecast that extends to a broader temporal horizon. }
    \label{fig:HQNN}
\end{figure*}

In this article, we present three types of hybrid quantum models as potential solutions for PV power forecasting. 
We assess the performance of our proposed models using a publicly accessible dataset~\cite{malvoni2016data}, encompassing a comprehensive array of meteorological variables as well as hourly mean PV power measurements spanning a $21$-month period. 
This dataset, along with the data preprocessing and analytical methods employed, is described in detail in Section~\ref{dataset}.

Our first proposed HQNN architecture, articulated in Section~\ref{HQNN}, incorporates classical fully connected layers with a vanilla variational repetitive quantum layer (VVRQ). 
Our second model, delineated in Section~\ref{HQNN2}, constitutes a hybrid quantum adaptation of the classical recurrent neural network, termed the Hybrid Quantum Long Short-Term Memory with quantum depth-infused layer (HQLSTM). 
While the first two models can predict the power for a certain hour ahead, the third model, presented in Section~\ref{HQNN3}, a Hybrid Quantum Sequence-to-Sequence Neural Network with quantum depth-infused layer, HQSeq2Seq, after training, is capable of forecasting PV power for arbitrary time intervals without requiring prior meteorological data. 

Despite using fewer trainable parameters, our hybrid quantum models outperform their classical counterparts in terms of more accurate predictions, including when trained on a reduced dataset. 
We summarize our conclusions and outline future research directions in Section~\ref{training ans results}.

\section{Materials and methods}\label{results}

The application of HQNNs in addressing time series prediction challenges, specifically in forecasting PV power output, offers several advantages. 
Primarily, their ability to operate within an exponentially larger computational search space enables them to efficiently capture intricate data patterns and relationships~\cite{kordzanganeh2023exponentially}. 
This feature not only enhances forecast accuracy~\cite{biamonte2017quantum, senokosov2024quantum} but also streamlines the learning process, requiring fewer iterations for model optimization~\cite{asel1}. Furthermore, the inherent capacity of quantum technologies to manage the uncertainty and noise ubiquitous in data offers more resilient and trustworthy predictions~\cite{schuld2018implementing}. 
This is particularly pertinent to power forecasting, given the inherent noise in meteorological data. 
Recent research also suggests that quantum models can be represented as partial Fourier series, positioning them as potential universal function approximators~\cite{schuld2021effect}, thereby broadening their applicability and efficacy in predictive tasks.

In terms of architecture, an HQNN is an amalgamation of classical and quantum components. 
The classical segments may consist of fully connected layers, convolutional layers, or recurrent layers, while the quantum segments are typically represented by variational quantum circuits (VQCs) or their contemporary modifications~\cite{kordzanganeh2023parallel, sedykh2024hybrid}.

The entire training workflow can be summarized as a five-stage pipeline, each stage being detailed in subsequent subsections:
\begin{enumerate}
    \item Data preprocessing.
    \begin{enumerate}
        \item A systematic cleaning pass is performed in which missing and inconsistent records are addressed. Gaps are imputed via piecewise–linear interpolation. An exploratory dependency analysis is conducted to identify dominant correlations among meteorological variables and PV output. Finally, all numeric features are rescaled to the unit interval $[0,1]$ using statistics computed only on the training partition to avoid target leakage; the same transformation is applied thereafter to validation and test splits.
        \item After being reduced to a common format, the series is transformed into a supervised learning format based on sliding windows tailored to the forecasting task. Each sample is defined by three parameters: the input window length (history used as context), the output window length (forecast horizon), and the stride by which the shift between consecutive windows is set. Multi–horizon prediction is supported, and temporal ordering preserved by chronological splits.
    \end{enumerate}
    \item Model development. Selection of the overall model architecture, including the choice between hybrid and classical configurations, the number of hidden layers, the types of layers (recurrent or feed-forward, quantum or classical), and the activation functions. 
    \item Analysis of quantum layers. For hybrid models, the embedded quantum circuit is analyzed to identify configurations that are able to maintain trainability and expressivity balance. Structural reducibility and over-parameterization are examined, and ranges for circuit hyperparameters are chosen to avoid barren–plateau behavior.
    \item Hyperparameter optimization and training. Hyperparameter optimization for both classical and hybrid models is performed using specialized frameworks, with a search space including learning rate, layer size, dropout rate, number of variational layers, and number of qubits. The optimal configuration is then trained using $k$-fold cross-validation.
    \item Model performance. Performance is assessed and then aggregated across folds using a comprehensive set of metrics, including MAE, MSE, RMSE, MAPE, $R^{2}$, and VAF. A final averaged estimate is provided together with a discussion of trade–offs between accuracy and efficiency.
\end{enumerate}

\begin{figure*}[ht]
    \includegraphics[width=1\linewidth]{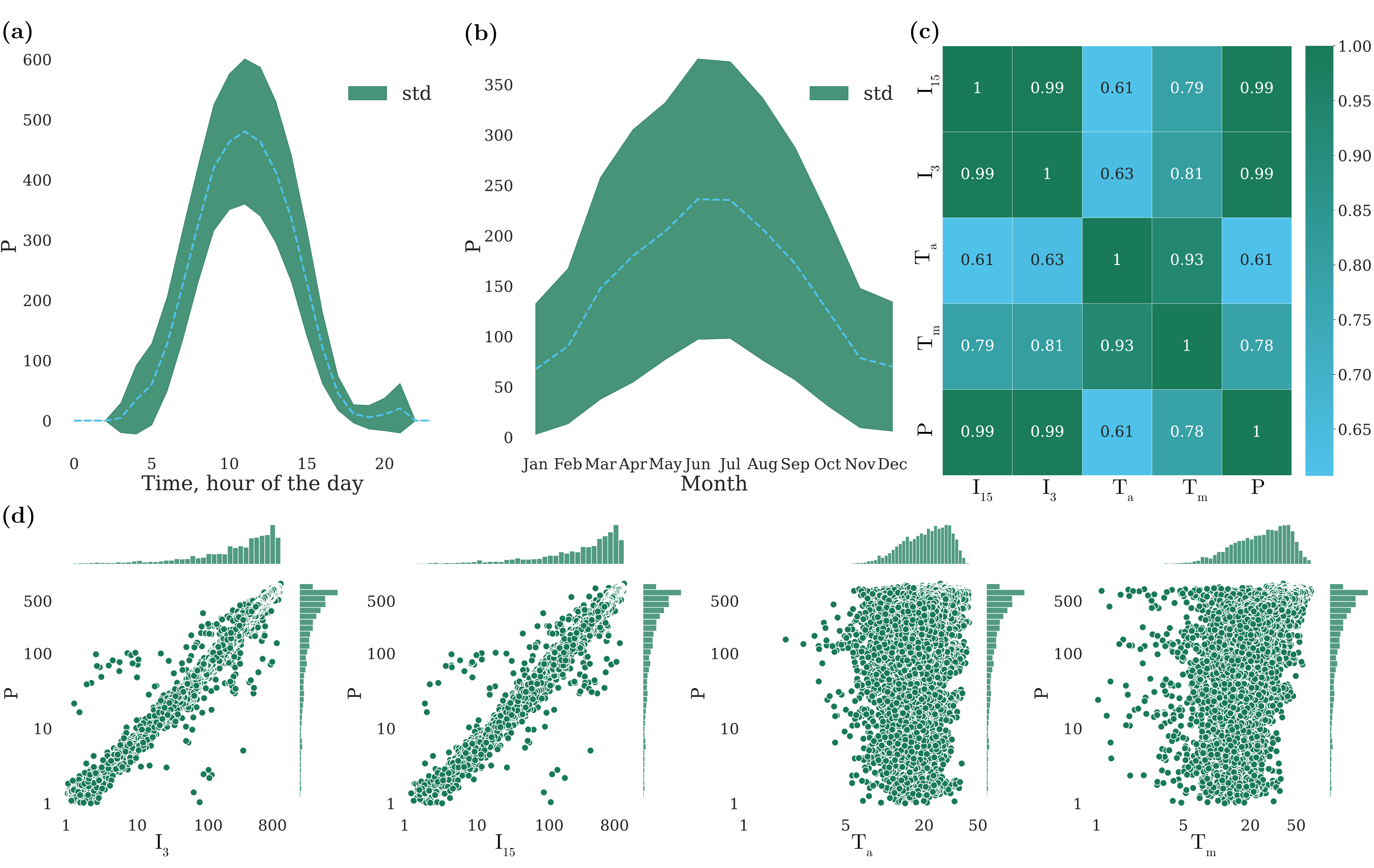}
    \caption{(a) Mean and standard deviation of the PV power value for each hour of the day. 
    (b) Mean and standard deviation of the PV power value for each month of the year. 
    The plot shows the PV power reaching maximal values in June and July. 
    (c) Correlation matrix of input features. 
    (d) Joint distribution of features.}
    \label{fig:data}
\end{figure*}

\subsection{Dataset}\label{dataset}

To investigate the advantages of hybrid quantum models, we selected a publicly accessible dataset~\cite{malvoni2016data} from a conventional generation plant situated in the Mediterranean region. 
This dataset is not only comprehensive but also enables benchmarking with results from various algorithms available in the literature. 
A comparative analysis of our model's predictions and those from the study by Ref.~\cite{kaloop2021novel} is provided in Section~\ref{training ans results}. 
The dataset, presented as a numerical table showcased in Fig.~\ref{fig:HQNN}, includes variables such as hourly mean ambient temperature, $T_a$; hourly mean module temperature, $T_m$; hourly mean solar irradiance recorded on two tilted planes with tilt angles of 3 and 15 degrees, $I_3$ and $I_{15}$; and hourly mean PV power, $P$, spanning $21$ months, accounting for more than $500$ days.

Beyond the scope of constructing models for predicting the output of PV panels, this dataset's utility extends to other applications. 
It aids in planning distributed battery energy storage systems~\cite{aghaei2019flexibility}, devising novel energy collection systems~\cite{kulkarni2018performance}, and researching the degradation patterns of PV panels~\cite{malvoni2017study}. 
The dataset's multifaceted applicability emphasizes its significance.


To ensure the validity and precision of the data, careful preprocessing was performed. 
We discovered approximately 20 anomalies in the original dataset. 
To maintain a continuous timeline, missing data points were replaced with the arithmetic mean of the preceding and succeeding days values. 
Additionally, data corresponding to the date ``12/31/13'' were excluded since it contained all-zero values, suggesting an error in the data collection. 
As a result, we obtained an uninterrupted dataset ranging from 4{:}00~AM on ``3/5/12'' to 12{:}00~AM on ``12/30/13.''

Additional in-depth analysis of the dataset was also conducted for a more nuanced understanding. 
Fig.~\ref{fig:data}(a) delineates the hourly distribution of PV power across the entire period. 
As expected, peak PV power values occur during midday, while nighttime values plummet to zero. 
Fig.~\ref{fig:data}(b) portrays monthly PV power fluctuations, which are more volatile compared to daily patterns, likely attributable to the limited number of full-year periods in the dataset. 
Fig.~\ref{fig:data}(c) presents a correlation matrix for the dataset features, identifying solar irradiances $I_3$ and $I_{15}$ as the features most correlated with PV power. 
Finally, the joint distribution of dataset features depicted in Fig.~\ref{fig:data}(d) further confirms that solar intensity is the feature most highly correlated with PV power.

\begin{figure*}[ht!]\label{HQLSTM}
    \includegraphics[width=1\linewidth]{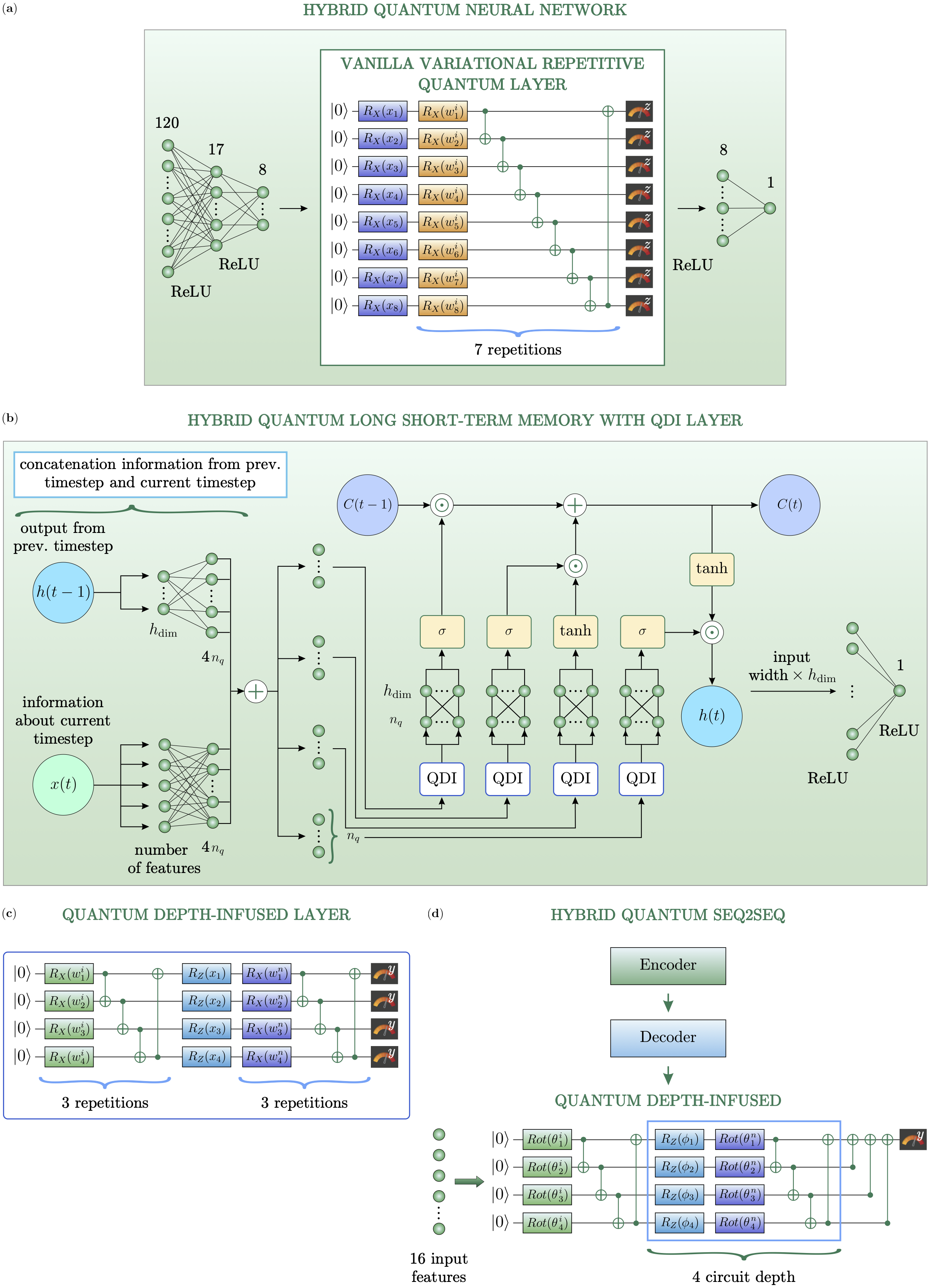}
    \caption{The architectures of: (a) Hybrid Quantum Neural Network with a VVRQ layer, 
    (b) Hybrid Quantum Long Short-Term Memory, 
    (c) QDI layer used in HQLSTM model, 
    (d) Hybrid Quantum Seq2Seq with QDI layer.}
    \label{fig:HQLSTM}
\end{figure*}

\subsection{HQNN}\label{HQNN}

This section introduces our first proposed model, referred to as the HQNN. 
As illustrated in Fig.~\ref{fig:HQNN}, the model accepts weather data spanning $24$ consecutive hours as its input. 
The output is a prediction of the PV power for the upcoming $25$th hour. 
The HQNN presented in Fig.~\ref{fig:HQLSTM}(a) is a combination of classical fully-connected layers, in our case with $120$, $17$, and $8$ neurons, and a VVRQ layer, which is a VQC, consisting of $q$ qubits and $d$ repetitions of variational layers, each distinguished by unique weights. 
The choice of $120$ neurons is methodical: the model ingests $5$ distinct features for each of the $24$ hours, resulting in a total of $120 = 5 \times 24$. 
The determination of the remaining parameters stemmed from an extensive hyperparameter optimization process, detailed in the subsequent sections.

Initially, every qubit in the VVRQ layer is set to the state $\ket{0}$. 
We subsequently encode the classical data by converting it into rotation angles around one of the $X$, $Y$, $Z$ axes using $R_x$, $R_y$, $R_z$ gates, respectively. 
This conversion employs the angle embedding technique~\cite{PhysRev.70.460}. 
For each qubit, the rotation angle, denoted by $x_j$, is determined by the $j$-th component of the input vector.

Following this, the variational layer is applied, which can either generate ``basic'' or ``strong'' entanglement. 
For the ``basic'' entanglement, each qubit undergoes a rotation by an angle $w_{j}^i$ around the $X$ axis,  followed by a layer of CNOT gates~\cite{Barenco_1995}. 
For the ``strong'' entanglement, each qubit is sequentially rotated by the angles $w^{(Z_1)}_{ji}$, $w^{(Y_2)}_{ji}$ and $w^{(Z_3)}_{ji}$ around the $Z$, $Y$ and $Z$ axes, respectively. 
This sequence is then followed by a layer of CNOT gates. 
In both cases, the variables $i$ and $j$ play crucial roles in determining the operations. 
The variable $i$ signifies the particular wire to which the operation is applied, and it takes values from the set ${1, 2, \dots, q}$. 
Meanwhile, the variable $j$ represents the number of variational layers and ranges from ${1, 2, \dots, d}$.

Lastly, all the qubits are measured in the Pauli-$Z$ basis, yielding the classical vector $v\in\mathbb{R}^q$.
This output serves as input for a subsequent classical fully-connected layer. 
This layer processes information from $q$ neurons into $1$ neuron that predicts the power value. 

The proposed HQNN model will be compared with its classical analogue -- a MLP that consists of $5$ fully connected layers with $120$, $32$, $3$, $3$, and $1$ neurons. 
The number of neurons in each layer was selected by a hyperparameter optimization procedure, detailed in the subsequent sections.

\subsection{HQLSTM}\label{HQNN2}

This section presents a description of our second hybrid model -- HQLSTM, which is a hybrid analogue of the classical LSTM model~\cite{hochreiter1997long}, with which predictions will be compared in the following sections. 
LSTM architectures have garnered significant attention in the realm of time series forecasting, including in predicting PV power~\cite{gao2023short, sharadga2020time, tovar2020pv}. 
HQLSTM models have performed well on tasks such as natural language processing~\cite{wang2022adverse}, detecting software vulnerabilities~\cite{akter2022automated}, and predicting solar radiation~\cite{yu2023prediction}.

In this proposed model, we added a quantum layer to each of the LSTM gates~\cite{chen2020quantum}. 
Our implementation is depicted in Fig.~\ref{fig:HQLSTM}(b). 
The input to the model is:
\begin{enumerate}
    \item The current step information, represented by a green circle, $x(t)$. 
    This is a tensor of size $5$, reflecting the five features for an hour, which include meteorological data and the PV power itself.
    \item The information from the previous step, denoted by a purple circle, $h(t-1)$. 
    It consists of a tensor of size $h_{dim}$. 
    For the initial step, this is simply a zero vector.
\end{enumerate} 
These inputs are processed through classical fully-connected layers to yield vectors with a uniform dimension of $4n_q$. 
These vectors are then combined via element-wise addition.

Subsequently, this concatenated vector is partitioned into four distinct groups for the four gates of the LSTM cell. 
As illustrated in Fig.~\ref{fig:HQLSTM}(b-c), each group is directed to the input of its corresponding quantum layer, symbolized by the Quantum Depth-Infused (QDI) square.

The outputs from QDI layers are transformed via classical fully-connected layers to standardize their dimensions to $h_{dim}$. 
Following this, similar to the classical LSTM, activation functions together with transformations appropriate to each of the four gates are applied to the outputs originating from the quantum layers. 
This processing produces the new cell state and hidden state vectors, $C(t)$ and $h(t)$, respectively.

The process operates in a cyclical manner. 
For each iteration, the vector from the current time step, $x(t)$, and the hidden vector from the previous step, $h(t-1)$, serve as inputs to the HQLSTM. 
This iterative process is executed as many times as the input width; in our case, the input width equals $24$. 
Subsequently, all the hidden vectors are concatenated to produce a single composite vector. 
This vector is then processed through a fully-connected layer consisting of a single neuron, which outputs a value that predicts the PV power.

In our first proposed architecture, the HQNN, the quantum layer functioned as a vanilla layer, where variational layers were sequentially placed after the encoding layer. 
In contrast, in the HQLSTM approach, we used a QDI layer~\cite{pharma} as depicted in Fig.~\ref{fig:HQLSTM}(c). 

To increase the expressivity, variational layers are positioned both before and after the encoding layer.
The first variational layer is illustrated with green rectangles, the encoding layer with blue rectangles, and the final variational layer with purple rectangles.

\subsection{HQSeq2Seq}\label{HQNN3}

Here we present a hybrid version of the Sequence-to-Sequence (Seq2Seq) model, first introduced in Ref.~\cite{sutskever2014sequence}. 
Seq2Seq models are widely used in natural language processing tasks~\cite{kalchbrenner2013recurrent}, where the length of input and output sequences is not predetermined and can be variable. 
We can also apply the principle of Seq2Seq models to the power prediction task~\cite{mu2023improved}. 
This means we can feed the neural network with time series of arbitrary length and produce a forecast for any number of hours ahead. 
In this problem setting, the longer the input time series is, the better the model prediction is. 
Similarly, the shorter the required output length, the easier it is for the model to generate the forecast.

The Seq2Seq model is a type of encoder-decoder model. 
The encoder is given the entire input sequence, which it uses to generate a context vector. 
This vector is used as an input hidden state for the decoder, so it literally provides it with ``context,'' according to which the decoder will generate the forecast. 
This means that the hidden dimensions of the encoder and the decoder must match.

The decoder creates the output sequence step by step. 
It starts with only the most recent entry.
Based on this entry and the context vector, the decoder generates the second entry and appends it to the existing one. 
The obtained two-entry sequence is once again fed into the decoder to generate the third entry. 
The cycle then repeats until the length of the generated sequence matches the length requested by the user.

We create and compare two models with Seq2Seq architecture: the classical Seq2Seq and the hybrid model called HQSeq2Seq. 
Both of these models have identical LSTMs acting as encoders and decoders. 
In the classical model, the decoder's hidden output vector is mapped to the ``Power'' value with a single linear layer, while in HQSeq2Seq it is processed by a QDI layer~\cite{pharma}.

In the QDI layer, instead of attempting to use a qubit for each feature~\cite{asel1}, we employed the data re-uploading technique~\cite{schuld2021effect, data_reuploading}. 
Specifically, we work with four qubits and structure them into a lattice of depth $4$ (depicted as a large blue rectangle in Fig.~\ref{fig:HQLSTM}(d)). 
Each of our $16$ input features leading to the quantum layer are intricately encoded within this lattice. 
The first four features are mapped onto the initial depth, followed by the subsequent features in blocks of four. 
Encoding these classical features into the quantum domain, we perform ``angle embedding'' using $R_z$ gates. 
This operation effectively translates the input vector into a quantum state that symbolizes the preceding classical layer's data. 

Entangling variational layers, signified by purple squares, are interspersed between every encoding layer, ensuring optimal Fourier accessibility. 
Each variational layer has two components: rotations governed by trainable parameters, and sequential CNOT gates. 
The rotations are implemented by quantum gates that manipulate the encoded input in line with the variational parameters, while the CNOT gates entangle the qubits. 

Each large blue bounding box encompasses a variational layer (small purple squares).
Prior to all the encoding layers, we introduce a single variational layer (small green squares) for enhanced model representation. 
Consequently, the total weight count in the quantum segment of our network is $20$. 
In the measurement phase, all the qubits execute a CNOT operation where the first qubit is the target. 
This ensures that when the first qubit is measured in the $Y$-basis, the measurement outcome depends on all the qubits involved in the circuit. 
In this way, the output of the quantum layer serves as the predicted power value for a specific hour.

The input size of the encoder and decoder can differ, which is a substantial benefit. 
For instance, we can use all of the $5$ features to create a context vector, but request to generate the forecast for only $1$ feature. 
Exploiting this advantage, we will feed the Seq2Seq model with a window of all known features and only demand the power forecast.
For the sake of simplicity, we trained both models with a fixed input and output length of $96$~hours and then vary the length in the testing stage.

\begin{figure*}[ht!]
    \includegraphics[width=1\linewidth]{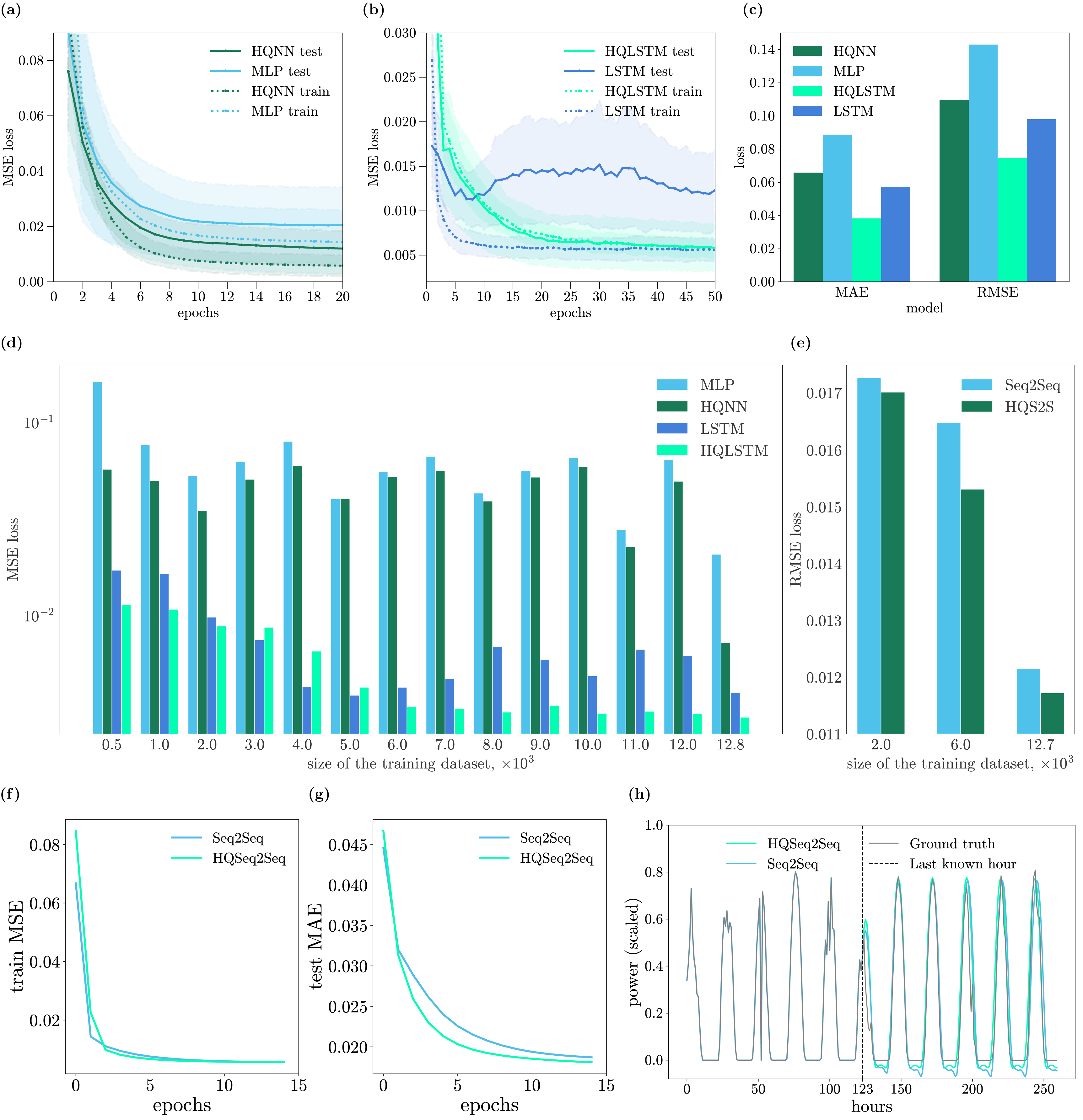}
    \caption{Results of training and testing for the HQNN, MLP, HQLSTM, and LSTM models. 
    (a-b) Training and testing history shown by the dotted and solid lines, respectively. 
    The filled region shows the standard deviation of the models averaged over the different testing subsets. 
    (c) Bar chart showing MAE and RMSE loss on the test dataset averaged over the results of 5 models trained on different subsets of training and test data.
    (d) The bar chart shows MSE loss on the test dataset for MLP, HQNN, LSTM and HQLSTM models for the reduced training dataset size.
    (e) The bar chart shows the RMSE loss on the test dataset for the Seq2Seq and HQSeq2Seq models for the reduced training dataset size.
    (f-g) The training and testing learning curves for the Seq2Seq and HQSeq2Seq models. 
    (h) Example for the classical and hybrid Seq2Seq models inference on the testing data. 
    The models receive $124$ hours of data as an input (before the dashed line) and forecast the ``Power'' up to $137$ hours ahead (after the dashed line). 
    The solid black line represents the ground-truth value of the ``Power'' feature. 
    The MSE/MAE errors in this particular example are $0.0052/0.0349$ for Seq2Seq and $0.0040/0.0292$ for HQSeq2Seq.
    \vspace{60pt}}
    \label{fig:results}
\end{figure*}

\section{Results and discussion}\label{training ans results}

In the study, six distinct models were employed for PV power prediction based on weather features: HQNN, MLP, HQLSTM, LSTM, HQSeq2Seq, and Seq2Seq. 
To train the models, the mean squared error (MSE) was chosen as the loss function: \[\text{MSE}= \frac{1}{N} \sum\limits_{n=1}^{N}\left(x_n-y_n\right)^2,\] where $N$ is the number of predictions, $\vec{x} = \left(x_1, x_2, \dots, x_N\right)$ is the predicted PV power, and $\vec{y} = \left(y_1, y_2, \dots, y_N\right)$ is the actual PV power value.

To test the models, in addition to the MSE loss metric, we also used the mean absolute error (MAE), root mean squared error (RMSE), variance accounted for (VAF), mean absolute percentage error (MAPE), and coefficient of determination ($R^2$).

The application of the squaring operation in the calculation of the RMSE results in the amplification of outlier influence, assigning significant weight to rare but substantial errors. Such errors can adversely affect power grid balancing processes and reserve capacity sizing. In contrast, the MAE and the MSE characterize the overall pointwise accuracy of a model and often demonstrate greater relevance for energy tasks. The VAF and the $R^2$ quantitatively assess the model's explanatory power by measuring the explained variance (higher values are preferable; $\text{VAF}=100\%$ or $R^2=1$ corresponds to a perfect fit). The MAPE provides a dimensionless measure of error in percentage terms, facilitating its operational interpretation; however, this metric may demonstrate instability when true values approach zero.

All the machine learning simulations for this study were conducted on CPUs using the QMware cloud~\cite{qmw_qmw_2022, kordzanganeh2023benchmarking} device. 
The classical part of our modeling was structured using the PyTorch library~\cite{PyTorch}, while the quantum part was implemented using the PennyLane framework. For our requirements, we selected the lightning.qubit device which is a custom backend for simulating quantum state-vector evolution. 
To compute the gradients of the loss function relative to each parameter: for the classical components of our hybrid models, we employed the widely-recognized backpropagation algorithm~\cite{backprop}; and for the quantum part, we used the adjoint method as highlighted in Refs.~\cite{Luo2020yaojlextensible, efficient_calculation}.

\begin{table}
\begin{tabular}{|ccc|}
\hline
\multicolumn{1}{|c|}{Hyperparameters}                                                                  & \multicolumn{1}{c|}{Range}                     & Best value            \\ \hline
\multicolumn{3}{|c|}{\textbf{HQNN}}                                                                                                                                             \\ \hline
\multicolumn{1}{|c|}{\begin{tabular}[c]{@{}c@{}}number of neurons in \\ the second layer\end{tabular}} & \multicolumn{1}{c|}{$8$\text{--}$128$}                 & 17                    \\ \hline
\multicolumn{1}{|c|}{number of qubits}                                                                 & \multicolumn{1}{c|}{$2\text{--}10$}                  & 8                     \\ \hline
\multicolumn{1}{|c|}{number of variational layers}                                                     & \multicolumn{1}{c|}{$1\text{--}10$}                  & 7                     \\ \hline
\multicolumn{1}{|c|}{embedding}                                                                        & \multicolumn{1}{c|}{$R_{x}$, $R_{y}$, $R_{z}$} & $R_{x}$               \\ \hline
\multicolumn{1}{|c|}{measurement}                                                                      & \multicolumn{1}{c|}{$X$, $Y$, $Z$}             & $Z$                   \\ \hline
\multicolumn{1}{|c|}{variational part}                                                                 & \multicolumn{1}{c|}{basic, strongly}           & basic                 \\ \hline
\multicolumn{1}{|c|}{initial learning rate}                                                            & \multicolumn{1}{c|}{$1\text{--}1000 \times 10^{-4}$} & $3 \times 10^{-2}$    \\ \hline
\multicolumn{3}{|c|}{\textbf{MLP}}                                                                                                                                              \\ \hline
\multicolumn{1}{|c|}{\begin{tabular}[c]{@{}c@{}}number of neurons in \\ the first layer\end{tabular}}  & \multicolumn{1}{c|}{$8\text{--}128$}                 & 32                    \\ \hline
\multicolumn{1}{|c|}{\begin{tabular}[c]{@{}c@{}}number of neurons in \\ the second layer\end{tabular}} & \multicolumn{1}{c|}{$8\text{--}128$}                 & 3                     \\ \hline
\multicolumn{1}{|c|}{\begin{tabular}[c]{@{}c@{}}number of neurons in \\ the third layer\end{tabular}}  & \multicolumn{1}{c|}{$8\text{--}128$}                 & 3                     \\ \hline
\multicolumn{1}{|c|}{initial learning rate}                                                            & \multicolumn{1}{c|}{$1\text{--}1000 \times 10^{-4}$} & $1 \times 10^{-2}$    \\ \hline
\multicolumn{3}{|c|}{\textbf{HQLSTM}}                                                                                                                                           \\ \hline
\multicolumn{1}{|c|}{\begin{tabular}[c]{@{}c@{}}number of neurons in \\ the second layer\end{tabular}} & \multicolumn{1}{c|}{$3\text{--}25$}                    & 20                    \\ \hline
\multicolumn{1}{|c|}{dropout range}                                                                    & \multicolumn{1}{c|}{$0\text{--}99$}                    & 0.239                 \\ \hline
\multicolumn{1}{|c|}{number of qubits}                                                                 & \multicolumn{1}{c|}{$2\text{--}5$}                     & 4                     \\ \hline
\multicolumn{1}{|c|}{number of variational layers}                                                     & \multicolumn{1}{c|}{$1\text{--}5$}                     & 1                     \\ \hline
\multicolumn{1}{|c|}{number of quantum layers}                                                         & \multicolumn{1}{c|}{$1\text{--}5$}                     & 3                     \\ \hline
\multicolumn{1}{|c|}{initial learning rate}                                                            & \multicolumn{1}{c|}{$1\text{--}100 \times 10^{-3}$}  & $0.52 \times 10^{-2}$ \\ \hline
\multicolumn{3}{|c|}{\textbf{LSTM}}                                                                                                                                             \\ \hline
\multicolumn{1}{|c|}{\begin{tabular}[c]{@{}c@{}}number of neurons in \\ the second layer\end{tabular}} & \multicolumn{1}{c|}{$3\text{--}25$}                    & 21                    \\ \hline
\multicolumn{1}{|c|}{dropout range}                                                                    & \multicolumn{1}{c|}{$0\text{--}99$}                    & 0.158                 \\ \hline
\multicolumn{1}{|c|}{initial learning rate}                                                            & \multicolumn{1}{c|}{$1\text{--}100 \times 10^{-3}$}  & $0.5 \times 10^{-2}$  \\ \hline
\end{tabular}
\caption{\label{tab:Hyperparameters} 
Summary of the hyperparameter optimization process for the HQNN, MLP, HQLSTM and LSTM models. 
It shows which hyperparameters are being optimized, the range over which they could be varied, and the best values found.}
\end{table}

\begin{table}[ht!]
\begin{tabular}{|c|c|c|c|c|c|}
\hline
Model      & HQNN & MLP & HQLSTM & LSTM   \\ \hline
train, MAE        &  $0.0458$     & $0.0651$       & \num[math-rm=\mathbf]{0.0382}      & $0.0454$           \\ \hline
test, MAE & $0.0659$         & $0.0887   $       &  \num[math-rm=\mathbf]{0.0343}      & $0.0570$             \\ \hline
train, MSE        &  $0.0059$      & $0.0144$       & $0.0056$       & \num[math-rm=\mathbf]{0.0054}                \\ \hline
test, MSE & $0.0120$         & $0.0204$       &  \num[math-rm=\mathbf]{0.0058}     & $0.0096$        \\ \hline
test, RMSE & $0.1097$         & $0.1428$       &   \num[math-rm=\mathbf]{0.0743}    & $0.0937$            \\ \hline
test, MAPE $\times 10^6$ & $33.5$         & $32.2$       &   \num[math-rm=\mathbf]{26.6}    & $40.8$            \\ \hline
test, VAF & $89.57$         & $89.71$       &   \num[math-rm=\mathbf]{92.01}    & $86.64$            \\ \hline
test, $R^2$ & $0.8703$         & $0.8767$       &   \num[math-rm=\mathbf]{0.9190}    & $0.8338$            \\ \hline
\end{tabular}
\caption{Summary of the results obtained by the HQNN, MLP, HQLSTM and LSTM models on the Mediterranean PV dataset~\cite{malvoni2016data}. 
}
\label{tab:summaryi}
\end{table}

\subsection{\textnormal{HQNN} versus \textnormal{MLP}, and \textnormal{HQLSTM} versus \textnormal{LSTM}}

Both the HQNN and its classical analogue, the MLP, were trained for $20$ epochs. 
In contrast, the HQLSTM and its classical counterpart, the LSTM, were trained for over $50$ epochs. 
The Adam optimizer~\cite{kingma2017adam} from the PyTorch framework was used to update the parameters of the models in order to minimize their loss functions. 
The comprehensive training process, accompanied by the results, is shown in Fig.~\ref{fig:results} and in Table~\ref{tab:summaryi}.

In this study, we employed cross-validation as a fundamental technique to assess the performance of our models across distinct testing subsets. 
The application of cross-validation is pivotal to safeguard against potential data leakage from the training dataset into the testing dataset. 
To achieve this, a rigorous approach was adopted in which a 24-hour time window on each side of the subsets was systematically excluded from the dataset.

Furthermore, we performed cross-validation by partitioning the dataset into training and testing sets in a $4:1$ ratio. 
This strategy was implemented to promote a comprehensive evaluation of our models, as we carried out model training and assessment on five distinct data splits. 
Subsequent averaging was used to consolidate the results obtained from these splits, and the mean values served as the primary metric for inter-model comparisons.

To achieve these results, we performed hyperparameter optimization using the Optuna optimizer~\cite{optuna_link}. 
The set of optimized parameters, the limits on their variation, and the best sets of hyperparameters for all our four models are presented in Table~\ref{tab:Hyperparameters}.

To verify the quality of the quantum layers assembled by Optuna, a theoretical analysis of quantum circuits was performed. Evaluation of ZX-calculus metrics, Fisher information, and Fourier series metrics showed that the resulting models possess high expressibility, trainability, and resource efficiency. The analysis is described in detail in Section~\ref{sec:appendix}.

After hyperparameter optimization, the total number of parameters for the models was $1109$, $2266$, $2857$, and $3987$ for the HQLSTM, HQNN, LSTM, and MLP models, respectively. 
The HQLSTM has the fewest parameters (approximately $39\%$ of its classical counterpart, the LSTM) and the HQNN has the second fewest parameters (approximately $57\%$ of its classical counterpart, the MLP). 

In a head-to-head comparison between the HQNN and the MLP, the former exhibits superior performance on 5 out of the 8 metrics. 
The largest training improvement was the $59\%$ reduction in the MSE, and the largest testing improvement was the $41\%$ MSE reduction (see Table~\ref{tab:summaryi}).

When comparing the HQLSTM and LSTM models, the HQLSTM was better on 7 out of the 8 metrics. 
Most significantly, both the test MAE and test MSE improved by about $40\%$. 
Moreover, the HQLSTM was more resistant to overfitting than the classical LSTM.

Regarding the two hybrid quantum models, the HQLSTM outperformed the HQNN in every metric. Its test MSE and test MAE scores were $52\%$ and $48\%$ better, respectively, while having half the number of trainable parameters.
In a broader comparison encompassing all four models, the HQLSTM emerged as the best model on 7 out of the 8 metrics. 
This is especially notable since the HQLSTM has the fewest trainable parameters of the four models.

Since the HQLSTM model demonstrates competitive performance, we further evaluated it against the established classical ensemble methods, XGBoost and LightGBM, to contextualize its effectiveness.
The HQLSTM has orders of magnitude fewer trainable parameters than these gradient-boosting tree algorithms.

As shown in Table~\ref{tab:cl_methods}, the HQLSTM model achieves the best values for 2 of the 7 metrics, RMSE and MAPE. 
There are negligible differences in the VAF and $R^2$ values, and the boosting models lead in MAE and MSE.

This is due to known differences in each metric's sensitivity: RMSE penalizes occasional large deviations more heavily, which are particularly disruptive for grid-balancing and reserve-capacity planning. Conversely, MAE and MSE capture overall point accuracy and may be preferable for revenue-settlement tasks. These findings highlight the importance of metric selection in model evaluation, with HQLSTM offering a compelling alternative for applications prioritizing error distribution robustness over absolute precision. 

It is also worth noting that the high value of the MAPE in all models is associated with a large number of zero values of the target variable during nighttime. Importantly, the HQLSTM reaches ensemble-level accuracy and does so with orders of magnitude fewer trainable parameters. Moreover, when we searched for external articles that refer to this dataset, we found only one article that solved the $1$ hour ahead PV power prediction using neural networks~\cite{kaloop2021novel}, and our HQLSTM model was better than the model from this article by $40\%$ ($92$ versus $65$) according to the VAF metric.

\begin{table}[ht!]\label{classical methods}
\begin{tabular}{|c|c|c|c|c|c|}
\hline
Model       & HQLSTM & XGBoost & LightGBM & Gain  \\ \hline
test, MAE          & $0.0266$       & $0.0260$       & \num[math-rm=\mathbf]{0.0251}   &  $3.46\%$         \\ \hline
test, MSE          & $0.0033$       & \num[math-rm=\mathbf]{0.0030}      & \num[math-rm=\mathbf]{0.0030}  &  $9.09\%$     \\ \hline
test, RMSE          & \num[math-rm=\mathbf]{0.0539}       & $0.0551$      & $0.0550$  & $2.00\%$     \\ \hline
test, MAPE $\times 10^6$ & \num[math-rm=\mathbf]{13.1}         & $20.7$       &   $14.4$  &  $9.03\%$ \\ \hline
test, VAF & $93.08$         & $93.75$       &   \num[math-rm=\mathbf]{93.77} & $0.02\%$ \\ \hline
test, $R^2$ & $0.9307$         & $0.9373$       &   \num[math-rm=\mathbf]{0.9375} & $0.02\%$ \\ \hline
\end{tabular}
\caption{Test performance comparison between the proposed HQLSTM model and classical ensemble methods such as XGBoost and LightGBM. 
For convenient interpretation of these results, the ``Gain'' column shows the relative improvement from the second best to the best value of the performance metric. The data was obtained for runs without k-fold validation.
}
\label{tab:cl_methods}
\end{table}

To further evaluate the generalization capability of HQLSTM beyond the original PV power prediction task, we conducted additional experiments using the Appliances Energy Prediction dataset~\cite{appliances_energy_prediction_374}. This dataset involves forecasting energy consumption in a smart home environment based on multiple contextual variables such as indoor temperature and humidity, outdoor weather conditions, and time-related features.

To ensure compatibility with our modeling pipeline, the original 10 minute resolution data were aggregated to hourly intervals and transformed into fixed-size time windows. Specifically, the model received 3 hours of past data to predict energy usage 1 hour ahead. Eleven features with the highest correlation to the target variable were retained for training.

\begin{figure*}[ht!]
    \includegraphics[width=1\linewidth]{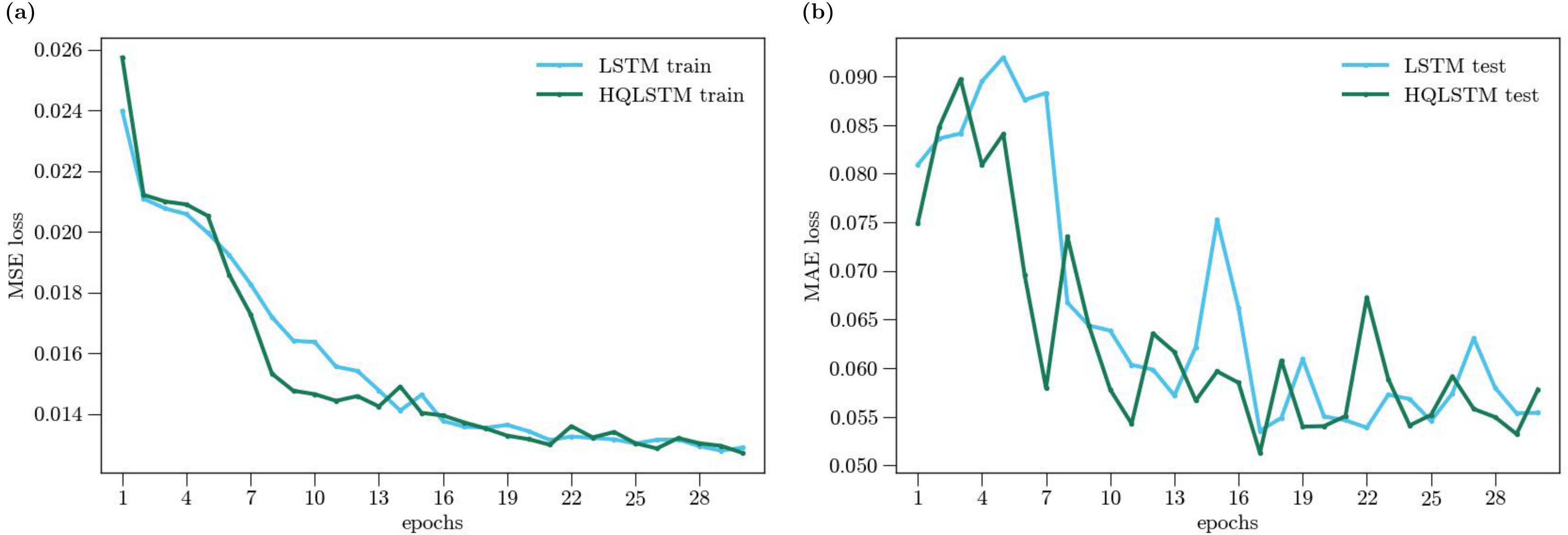}
    \caption{Training dynamics of HQLSTM and LSTM models on the energy prediction task. (a) MSE loss on the training subset shows both models converging similarly. (b) MAE loss on the test subset indicates that HQLSTM achieves lower test loss values overall. Specifically, the lowest loss is $0.0513$ and $0.0535$ for the HQLSTM and LSTM models, respectively.}
    \label{fig:results_add_data}
\end{figure*}

Both classical LSTM and HQLSTM models were retrained under the new settings with the HQLSTM having $20622$ trainable parameters and the LSTM having $89257$. 
A head-to-head comparison is shown in Table~\ref{tab:additional_data} and Fig.~\ref{fig:results_add_data}, where it can be seen that the HQLSTM excels in 3 out of the 6 performance metrics.
The largest relative improvement is the $12.5\%$ reduction in the MAPE, which shows that the HQLSTM has better control over its typical and relative errors.

\begin{table}[ht!]
\begin{tabular}{|c|c|c|c|c|}
\hline
Model      & HQLSTM & LSTM & Gain \\ \hline
test, MAE &   \num[math-rm=\mathbf]{0.0513}      & $0.0535$ &  $4.11\%$          \\ \hline
test, MSE &  $0.0090$     & \num[math-rm=\mathbf]{0.0089} &  $1.11\%$     \\ \hline
test, RMSE &  $0.0937$     & \num[math-rm=\mathbf]{0.0934} & $0.32\%$      \\ \hline
test, MAPE &  \num[math-rm=\mathbf]{38.39}     & $43.88$ & $12.51\%$      \\ \hline
test, VAF &  \num[math-rm=\mathbf]{38.07}     & $37.83$  & $0.63\%$     \\ \hline
test, $R^2$ &  $0.3595$     & \num[math-rm=\mathbf]{0.3682} & $2.42\%$      \\ \hline
\end{tabular}
\caption{Performance comparison between HQLSTM and LSTM on the additional energy consumption dataset. 
As before, ``Gain'' is the relative improvement from the second best to the best value of the performance metric.}

\label{tab:additional_data}
\end{table}

These results suggest that HQLSTM is more robust to individual prediction errors, despite using $4$ times fewer trainable parameters. They also confirm that the proposed HQLSTM architecture generalizes well to new forecasting tasks and retains strong performance characteristics across domains.

Further, to confirm that hybrid models train better, including on a smaller dataset, an additional experiment was conducted where the volume of training data was intentionally reduced. 
The results are shown in Figure~\ref{fig:results}(d–e). 
The hybrid models performed better with less data, having lower losses and better prediction capabilities than the classical models. 

\subsection{\textnormal{HQSeq2Seq} versus \textnormal{Seq2Seq}}

After the preprocessing steps described in the Section~\ref{dataset}, the dataset spanned $12775$ hours from 3/5/12 4:55 AM to 8/19/13 10:00 AM for training, and $3194$ hours from 8/19/13 11:00 PM to 12/30/13 00:00 AM for testing.

Although the models are capable of being trained on sequences of arbitrary lengths, we chose to use sequences with a fixed $96$~hours length for simplicity. 
In this case, the encoder gets $96$~hours of all available features, while the decoder is asked to extrapolate only the ``Power" feature of the data $96$~hours ahead. 
The Seq2Seq model had 2705 trainable parameters and the HQSeq2Seq model had 2750.

Model training was conducted using the Adam optimizer (learning rate $0.001$) with a maximum of 15 epochs. 
We additionally applied early stopping based on test loss, and observed that both models typically converged within the first $13\text{--}15$ epochs (see Fig.~\ref{fig:results}(f–g)). 
As such, extending training further provided no meaningful performance gains, while increasing computational cost.

As an example of inference, we pass the time series of a length different from $96$ into both models and prompt them to give us a forecast for $137$ hours ahead (Fig.~\ref{fig:results}(h)). 
Both models generalize well from the fixed sequence length to an arbitrary one. 
It may even be possible to improve these results by introducing variable-length sequences into the training stage.

We also investigated the dependence of the test loss on the size of the training dataset for Seq2Seq and HQSeq2Seq, as shown in Fig.~\ref{fig:results}(e). 
As one can see, the test RMSE loss of the hybrid model is lower for any size of training data, showing that the hybrid model has an advantage over the classical model, including on a substantially trimmed dataset.

\subsection{Discussion}\label{plans}

In this work, we introduced three hybrid quantum approaches to time series prediction. 
The first two models allow one to predict the power of solar panels $1$ hour ahead using weather features from the previous $24$ hours. 
The third model is the most versatile since it enables longer-term, user-defined forecasts to be made.

The first approach is the HQNN, a combination of classical fully connected layers and a VVRQ layer, which is a quantum analogue of such layers. 
We compared this hybrid model with its classical counterpart, the MLP, and demonstrated that, even though the HQNN has $1.8$ times fewer variational parameters, its predictive ability is $41\%$ better, as estimated by the MSE metric.

The second approach is the HQLSTM, a hybrid quantum analogue of the classical LSTM. 
Here, a QDI layer is inserted into each gate of the LSTM cell. 
This approach provides a $40\%$ improvement in prediction using the MAE and MSE metrics compared to its classical counterpart. 
The HQLSTM was even an improvement on the HQNN despite having half as many weights.

The third approach is the hybrid Seq2Seq model, which consists of two LSTMs with a quantum layer at the end. 
This approach allows one to predict the PV power not only for an hour ahead, but for any number of hours ahead, without knowing the weather features in advance. 
The addition of the proposed QDI layer improves the accuracy of the predictions, reducing the 
MAE error by $16\%$ compared to a purely classical Seq2Seq model.

We conducted an additional experiment in which all our models were trained on a reduced dataset.
It confirmed that hybrid models have better learning capabilities and lower losses than their classical counterparts when trained on any amount of data. 
This motivates the use of hybrid networks for applications where data collection is a complex task.

We also compared our models with those reported in a paper that solved the same problem using the same dataset. 
We found that our best HQLSTM is $40\%$ more accurate in predicting power using the VAF metric.

To complement our performance evaluation, we conducted an architectural analysis of the quantum layers used in our models. As detailed in Section~\ref{sec:appendix}, we compared the VVRQ and QDI layers using ZX-calculus, Fisher information matrices, and Fourier spectral analysis. These methods allowed us to assess the efficiency, trainability, and expressive capacity of the circuits beyond their end-task performance. The results reveal concrete trade-offs. VVRQ is ZX-irreducible (all $56/56$ parameters retained) and shows near-uniform FIM spectra without rank saturation as depth grows, indicating depth-scalable trainability; its Fourier analysis (approximate, due to size) indicates an increasing number of non-zero coefficients with qubit count, consistent with broad expressivity. In contrast, QDI is also ZX-irreducible ($24/24$ parameters) but achieves high trainability at shallow depth: its FIM enters the over-parameterized regime (rank saturation), while its Fourier spectrum is densely populated ($151/161 \approx  94\%$ non-zero coefficients), making it compact and well-suited for resource-constrained hybrids such as HQLSTM. Taken together, these measurements clarify the design space of quantum components and support informed choices between depth-scalable VVRQ and lightweight yet expressive QDI in future hybrid QML architectures.

To fully unlock the potential of hybrid quantum approaches in time series prediction problems, further research and testing of models on other datasets are needed. 
More efficient optimization and training of VQCs, combined with higher fidelity and larger-scale quantum hardware, could lead to significantly greater performance improvements. 
While the study focuses on performance characteristics of hybrid quantum-classical models, we acknowledge that their training and tuning can be more complex than those of purely classical counterparts. This is due to the need for domain knowledge in both machine learning and quantum computing, as well as simulator-related constraints. Future work may explore more accessible training workflows and improved tool-kits for hybrid model development.

While this work was done on a public dataset with an emphasis on hybrid quantum models for better forecasting performance, the quality and source of data play a crucial role in overall effectiveness in the real world, especially considering weather data. 
Accurate weather forecasts are a crucial input to any high performing and useful PV prediction given its dynamism and influence on PV output. 
An interesting area of research is cloud prediction using satellite and weather data for geolocations directly impacting solar irradiance and therefore PV output. 
The added complexity could enhance the need for hybrid quantum models to increase computational efficiency and to provide higher quality forecasts.

\subsection{Conclusion}

In this work, we addressed the important problem of photovoltaic time-series forecasting. 
This is a crucial task for successfully operating solar power stations and managing their integration with the electricity grid. 
To this end, we proposed three hybrid quantum–classical time-series models, namely HQNN, HQLSTM and HQSeq2Seq, and evaluated them on a dataset from a Mediterranean power plant. 
We used five-fold cross-validation to carefully analyze their performance relative to their most comparable classical counterparts. 
We made additional comparisons with XGBoost and LightGBM, performed tests where we deliberately reduced the size of the training data, and assessed whether the HQLSTM model could generalize to an energy-consumption prediction task. 
Our contributions can be divided into three parts. First, we integrated quantum circuits (VVRQ and QDI) into standard MLP, LSTM, and Seq2Seq backbones. Second, we developed an HQSeq2Seq model that can forecast user-defined horizons without requiring future meteorological inputs. Third, we included a circuit-level analysis based on ZX-calculus, Fisher information, and Fourier metrics. 

The main outcomes of the study were as follows. 
The HQLSTM, with only 1109 parameters, achieved a test MAE of 0.0343 and a test MSE of 0.0058, while the LSTM, with 2857 parameters, had scores of 0.0570 and 0.0096, respectively.
The HQSeq2Seq reduced the MAE by 16$\%$ compared to the classical Seq2Seq.
The hybrid models retained their advantages when the training set was substantially reduced.
The HQLSTM reached ensemble-level performance on RMSE and MAPE with far fewer parameters.
This model generalized to a different energy dataset with lower typical errors while using about 4 times fewer trainable parameters.

Future work could include validating the models across additional locations and with higher-frequency settings, developing more efficient and user-accessible approaches to the training of variational quantum circuits, and evaluating performance on actual (rather than simulated) quantum devices.\\

{\bf Declaration of competing interest }

The authors declare that they have no known competing financial 
interests or personal relationships that could have appeared to influence 
the work reported in this paper. \\

{\bf Data availability }

All datasets are publicly available and are cited within the Dataset subsection of the paper. The machine learning models employed are fully described in the Results section, enabling independent reproduction of the results.

\bibliography{qml_hybrid_pv}
\bibliographystyle{unsrt}

\section{Supplementary Material}\label{sec:appendix}

In this section, we conduct a detailed analysis of the quantum layers used in the HQNN and HQLSTM architectures, namely the VVRQ and QDI circuits. While isolating their exact contribution to model performance is non-trivial due to the hybrid and entangled nature of training, we instead evaluate their design from three complementary perspectives:
\begin{itemize}
\item ZX calculus for reducibility
\item Fisher information for trainability
\item Fourier analysis for expressivity
\end{itemize}

This analysis allows us to better understand the architectural properties of the quantum layers and their relative strengths in terms of efficiency, trainability, and representational power.

\begin{figure*}[ht]
    \centering
    \includegraphics[width=1\linewidth]{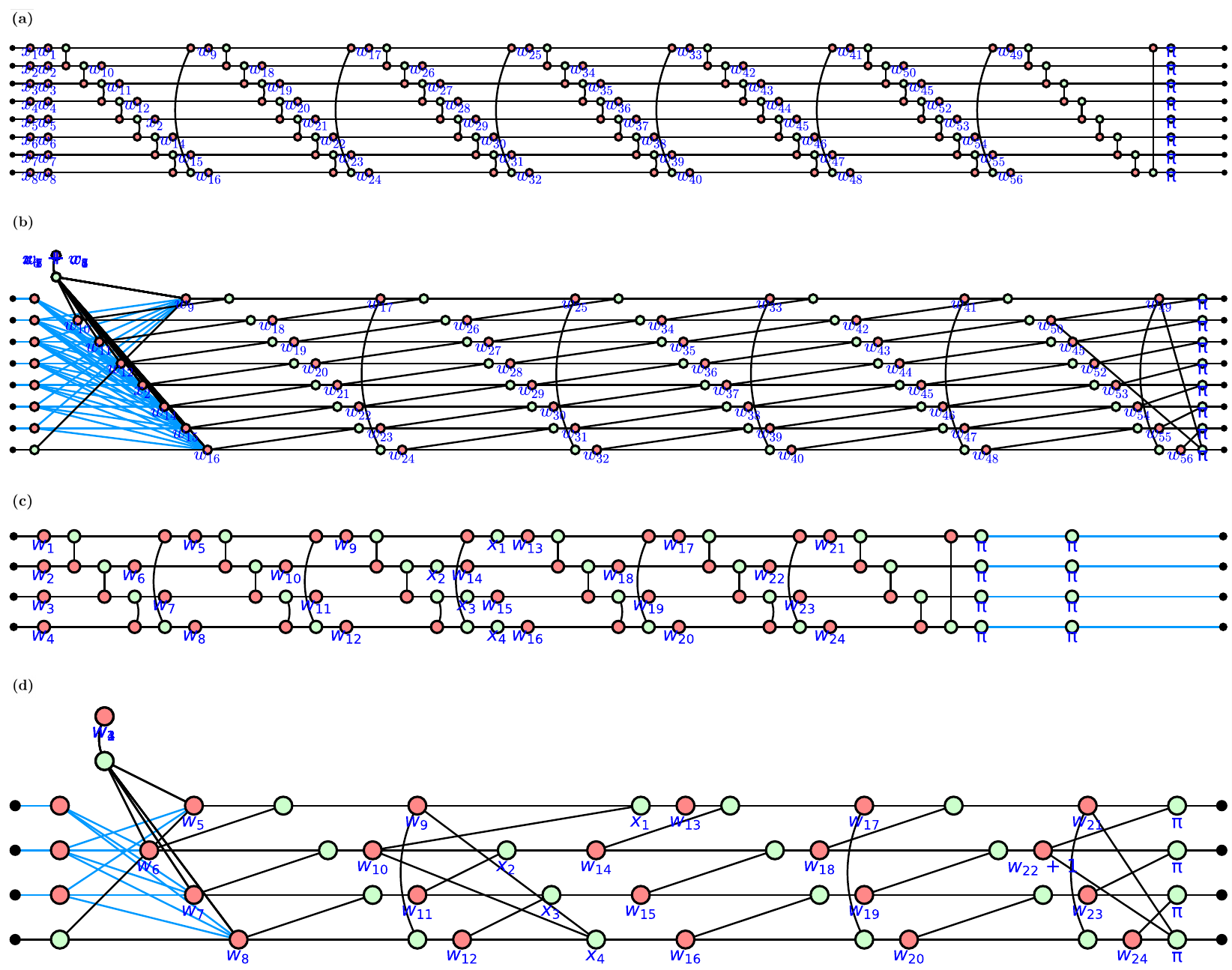}
\caption{\textbf{(a)} VVRQ layer: original configuration. \textbf{(b)} VVRQ layer: optimized configuration. \textbf{(c)} QDI layer: original configuration. \textbf{(d)} QDI layer: optimized configuration}
\label{fig:ZX}
\end{figure*}

\subsection{ZX calculus}\label{sec:appendix_ZX}

ZX-calculus provides a diagrammatic language in which quantum circuits are represented by ``spiders'' (nodes) connected by edges. These ZX diagrams can be systematically simplified via graphical rewriting rules grounded in the nature of quantum mechanics~\cite{Coecke2008ZX, wetering2020zx}. The resulting simplification yields a more resource-efficient circuit and offers a quantitative measure of efficiency (reducibility): the ratio between the parameter count in the reduced diagram and that in the original circuit. A circuit for which no further simplification is possible is regarded as ZX-irreducible.

Applying this methodology to the circuit in Fig. \ref{fig:ZX}(a–b) shows that all $56$ of the initial $56$ parameters (i.e., $100\%$) are retained after reduction, confirming that the VVRQ circuit is already in an optimal form. A similar outcome is obtained for the QDI layer in Fig. \ref{fig:ZX}(c–d): the ZX-calculus procedure leaves all $24$ of the original $24$ parameters intact, again indicating a fully optimized, ZX-irreducible circuit. The QDI layer preserves all $24$ parameters after simplification, indicating a minimal and expressive configuration with no redundant structure.

\subsection{Fisher information}\label{sec:appendix_fim}

These results suggest that while VVRQ is more scalable in depth before saturation, QDI is inherently compact and achieves excellent trainability with minimal layers.

A supervised learning task entails constructing a parametric hypothesis
\(h_{\theta}(\hat{x})\) from a labeled data set \((x,y)\in X\times Y\) that
approximates the underlying data-generating distribution \(f(x)\).
Given a subset of \(S\) labeled examples, the parameters \(\theta\) are
optimized to maximize the likelihood that the model assigns the correct
label \(y\) to an input \(x\); the relevant conditional likelihood is
\(P(y\mid x,\theta)\).
Assuming a uniform prior over \(X\), it is convenient to work with the joint
probability \(P(y,x\mid\theta)\), which can be evaluated for any \(\theta\)
at each sample \(x_i\).
The collection of such probabilities forms a differentiable
manifold of dimension \(N=|\theta|\), where \(N\) is the number of
trainable parameters.

The geometry of this manifold is captured by the Fisher information
matrix (FIM)~\cite{abbas2020power,amari1998gradient},

\begin{equation}
    F(\theta)=\mathbb{E}_{\{x_i,y_i\}}
    \bigl[\nabla_{\theta}\log P\,\nabla_{\theta}\log P^{\mathsf T}\bigr].
\end{equation}

Diagonalizing \(F(\theta)\) yields a locally Euclidean basis whose
eigenvalues correspond to squared gradient magnitudes.  
Their distribution is critical for diagnosing the barren-plateau phenomenon, where gradients vanish as system size increases~\cite{mcclean2018barren}.
Sampling the FIM spectrum across many parameter realizations therefore
provides an effective metric of trainability -- metric of how weights in the model are contributing to the output (not just zero or not as in ZX-calculus).

Following Abbas \emph{et al.}~\cite{abbas2020power}, we evaluate the FIM for the concrete hyper-parameters of our circuits using a synthetic Gaussian dataset \(x\sim\mathcal{N}(\mu=0,\sigma^{2}=1)\).  
The joint probability is obtained from the overlap between the computed quantum state and the state prepared by the circuit,
\begin{equation}
    P(y,x\mid\theta)=\langle y|\psi(\theta,x)\rangle,
\end{equation}
and is averaged over all \(x\) and \(y\) to obtain \(F(\theta)\) for each
\(\theta\).

\begin{figure*}[!ht]
    \centering
    \includegraphics[width=1\linewidth]{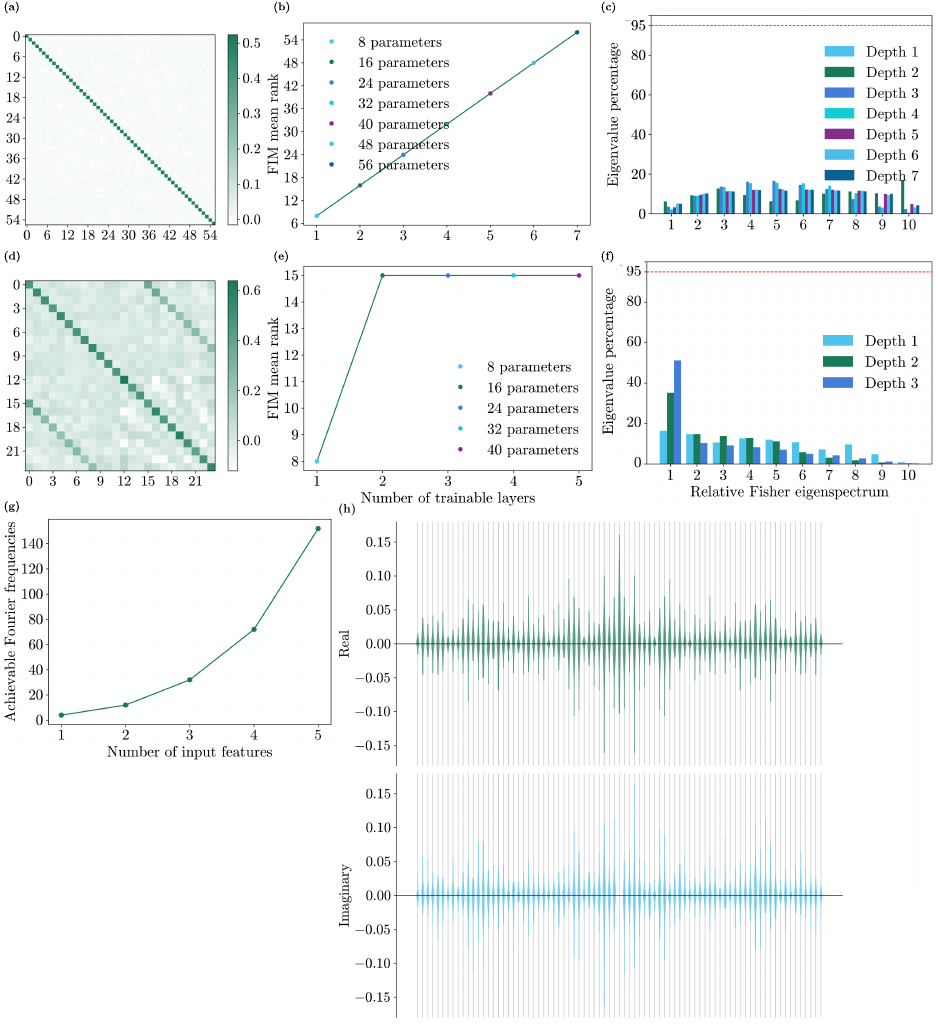}
\caption{\textbf{(a)}
Mean Fisher information matrices for VVRQ. The diagonal elements confirm an even
distribution of gradients across all trainable parameters, with no
single-parameter dominance. \textbf{(b)} FIM rank for VVRQ ($7/7$ layers used) illustrates the circuit's limit of over-parameterization. Although VVRQ could in principle benefit from additional layers, practical hardware constraints limit further scaling. \textbf{(c)} The normalized histogram of the VVRQ Fisher eigenspectrum. For VVRQ, all seven layers show good parameter distribution, although the $7$th layer provides the most uniform distribution. This circuit thus has very good trainability and is robust against the barren plateau problem.
\textbf{(d)}
Mean Fisher information matrices for QDI. QDI exhibits a similar pattern as the VVRQ (aside from a slight shift toward later parameters in deeper layers) indicating high
trainability. \textbf{(e)} FIM rank for QDI ($3/3$ layers used) illustrates the circuit's limit of over-parameterization. QDI is almost optimally compact for integration in a HQLSTM, with only one extra layer above the saturation point --- acceptable trade-off for increased expressivity. \textbf{(f)} The normalized histogram of the QDI Fisher eigenspectrum. For QDI, the majority of parameters also achieved distribution very close to uniform with modest number near-zero eigenvalues, which indicates excellent training capabilities.
\textbf{(g)}
VVRQ layer's Fourier spectrum data for a smaller number of qubits. \textbf{(h)} QDI layer's Fourier spectrum.}
\label{fig:fisher}
\end{figure*}

Figure~\ref{fig:fisher}(a), (d) confirms that every parameter contributes
substantially: in VVRQ the contributions are almost identical, as well as in QDI.

Larocca \emph{et al.}~\cite{larocca2021overparam} showed that certain QNNs
become over-parameterized when the FIM rank ceases to grow with added
layers. As illustrated in Fig.~\ref{fig:fisher}(b), (e), only QDI has reached this saturation. VVRQ thus remains expressivity-limited primarily by hardware resources, whereas QDI can utilize the full advantages of the over-parameterization regime.

Finally, the relative eigenspectra in Fig.~\ref{fig:fisher}(c)-(f) exhibit an almost uniform distribution for the VVRQ architecture and similarly for most of the QDI parameters, indicating a strong immunity to barren plateaus and hence excellent trainability.

\subsection{Fourier series}\label{sec:appendix_fourier}

It is well established that the output of a parameterized quantum circuit can
be written as a truncated Fourier series~\cite{schuld2021encoding,Schuld2023Fourier,Parfait2024Fourier}. For an
input vector \(x\in\mathbb{R}^{N}\) and trainable parameters
\(\theta\), the model output takes the form

\begin{equation}
    f_{\theta}(x)= \sum_{\omega_1\in\Omega_1}\cdots\sum_{\omega_N\in\Omega_N}
    c_{\omega_1\ldots\omega_N}(\theta)\,e^{-i\omega x},
\end{equation}
where each frequency component \(\omega_i\) lies in
\(\{-d_i,\ldots,0,\ldots,d_i\}\).  Hence the number of Fourier terms along
dimension \(i\) is \(2d_i+1\), with \(d_i\) equal to the number of times
that feature is encoded in the circuit~\cite{perez2020data}.

This allows us to introduce expressivity --- metric that shows the complexity of patterns in data which our circuit can learn. To characterize the expressive power of \(f_{\theta}(x)\) we draw each \(\theta_i\) independently from the uniform interval
\([0,2\pi]\) and sample \(x\)-values on an equidistant grid whose frequency matches \(d\).  

Two aspects are of primary
interest:\\
\textbf{(a)} Degree: the total number of non-zero Fourier terms\\
\textbf{(b)} Coefficient accessibility: the extent to which the trainable
weights can populate those terms.\\
For each random realization of
\(\theta\) we compute the complex coefficients
\(c_{\omega_x,\omega_y}(\theta)\) and visualise their distributions as
violin plots.  A highly expressive circuit exhibits many coefficients that
deviate significantly from zero in both their real and imaginary parts; a
non-expressive circuit yields coefficients tightly clustered near zero. The spread of these distributions is quantified by the sample standard
deviation across \(\theta\).

The end result, as can be seen in Fig.\ref{fig:fisher}(g)-(h), is the Fourier spectrum violin plot. For VVRQ, as it is too big to calculate it upfront, an approximation was used to indicate the number of non-zero coefficients. It grows with the number of qubits, showing good expressivity for several previous configurations. As for QDI, $151$ out of $161$ (around $94\%$) coefficients aren't zero, which shows an excellent expressive power of this architecture. The broader and more populated Fourier spectrum for VVRQ indicates its potential to model more complex functions, while QDI retains a focused yet sufficiently expressive frequency support.

\end{document}